\documentclass[a4paper]{article}

\usepackage{INTERSPEECH2019}
\usepackage{times}
\usepackage{latexsym}
\usepackage{multirow}
\usepackage{amsmath}
\usepackage{hyperref}

\newcommand{\src}{\boldsymbol s}
\newcommand{\trg}{\boldsymbol t}
\newcommand{\lsrc}{{|\boldsymbol s|}}
\newcommand{\ltrg}{{|\boldsymbol t|}}
\newcommand{\R}{\mathbb{R}}
\DeclareMathOperator\softmax{softmax}

\def\bea{\begin{eqnarray}}
\def\eea{\end{eqnarray}}

\title{Empirical Evaluation of Sequence-to-Sequence Models for Word Discovery in Low-resource Settings} 

\name{Marcely Zanon Boito$^1$, Aline Villavicencio$^{2,3}$, Laurent Besacier$^1$}
\address{
  $^1$Laboratoire d'Informatique de Grenoble, Univ. Grenoble Alpes (UGA), France\\
  $^2$School of Computer Science and Electronic Engineering, University of Essex, UK\\
  $^3$Institute of Informatics, Federal University of Rio Grande do Sul, Brazil}
\email{contact: marcely.zanon-boito@univ-grenoble-alpes.fr}

\begin{document}
\maketitle
\begin{abstract} 
Since Bahdanau et al.~\cite{bahdanau2014neural} first introduced attention for neural machine translation, most sequence-to-sequence models made use of attention mechanisms~\cite{elbayad2018pervasive, vaswani2017attention, gehring2017convolutional}. While they produce soft-alignment matrices that could be interpreted as alignment between target and source languages, we lack metrics to quantify their quality, being 
unclear which approach produces the best alignments.
This paper presents 
an empirical evaluation of 3 of the main sequence-to-sequence models for word discovery from unsegmented phoneme sequences: CNN, RNN and Transformer-based.  This task consists in aligning word sequences in a source language with 
phoneme sequences in a target language, inferring 
from it 
word segmentation on the target side~\cite{godard2018unsupervised}.
Evaluating word segmentation quality can be seen as an extrinsic evaluation of the soft-alignment matrices produced during 
training.
Our experiments in a low-resource scenario on Mboshi and English languages 
(both aligned to French) show that RNNs surprisingly outperform CNNs and Transformer for this task. 
Our results are confirmed by an intrinsic evaluation of 
alignment quality through the use 
Average Normalized Entropy~(ANE). 
Lastly, we improve our best word discovery model by using an alignment entropy confidence measure that accumulates ANE over all the occurrences of a given alignment pair in the collection.
\end{abstract}
\noindent\textbf{Index Terms}: sequence-to-sequence models, soft-alignment matrices, word discovery, low-resource languages, computational language documentation

\section{Introduction}

Sequence-to-Sequence~(S2S) models can solve many tasks where source and target sequences have different lengths. For learning to focus on specific parts of the input at decoding time, most of these models are equipped with attention mechanisms~\cite{bahdanau2014neural,elbayad2018pervasive, vaswani2017attention, gehring2017convolutional, NIPS2014_5346}.
By-products of the attention are soft-alignment probability matrices, that can be interpreted as alignment between target and source. However, 
we lack metrics to quantify their quality. 
Moreover, while these models perform very well in a typical use case, it is not clear how they
would be affected by low-resource scenarios. 

This paper proposes an empirical evaluation of well-known 
S2S models for a particular S2S  
modeling task. This task 
consists of aligning word sequences in a source language with phoneme sequences in a target language, inferring 
from it 
word segmentation on the target side~\cite{godard2018unsupervised}. 
We concentrate on three 
models: Convolutional Neural Networks (CNN) \cite{elbayad2018pervasive},  Recurrent Neural Networks (RNN) \cite{bahdanau2014neural} and 
Transformer-based models \cite{vaswani2017attention}. 
While this \textit{word segmentation} task can be used for the extrinsic evaluation of the soft-alignment probability matrices produced during S2S 
learning, we 
also introduce Average Normalized Entropy (ANE), a task-agnostic confidence metric 
to quantify the quality of the  source-to-target alignments obtained.
Experiments performed on a low-resource scenario for two languages (Mboshi and English) using equivalently sized corpora aligned to French, are, to our knowledge, the first empirical evaluation of 
these well-known S2S 
models 
for a  word segmentation task. 
We also illustrate how our entropy-based metric can be used in a language documentation scenario, helping a linguist to efficiently discover types, in an unknown language, from an unsegmented sequence of phonemes.
This work is thus also a contribution to the emerging 
computational 
language documentation domain~\cite{adda2016breaking,anastasopoulos2017case,besacier2006towards, lignos2010recession,bartels2016toward}, whose main goal is the creation of automatic approaches able to help the documentation of the many languages soon to be extinct~\cite{austin2011cambridge}.

Lastly, studies focused on comprehensive attention mechanisms for NMT~\cite{song2018hybrid, li2018multi, voita2019analyzing} lack evaluation of the resulting alignments, and the exceptions~\cite{zenkel2019adding} do so for the task of word-to-word alignment in well-resourced languages. 
Differently, our work is not only an empirical evaluation of NMT 
models focused on alignment quality, but it also tackles data scarcity of low-resource scenarios. 


\section{Experimental Settings}
\label{sec:settings}

\subsection{Unsupervised Word Segmentation from Speech}\label{AWS}

As in language documentation scenarios available corpora usually contain speech in the language to document aligned with translations in a well-resourced language, Godard et al.~\cite{godard2018unsupervised} introduced a pipeline for performing Unsupervised Word Segmentation~(UWS) from speech. The system 
outputs time-stamps delimiting stretches of speech, associated with class labels, corresponding to real words in the language. The pipeline consists of first transforming  speech into a sequence of phonemes, 
either through 
Automatic Unit Discovery~(e.g.~\cite{ondel2016variational}) or 
manual transcription. The phoneme sequences, together with their translations, are then fed to an attention-based S2S 
system that produces soft-alignment probability matrices between target and source languages. The alignment probability distributions between the phonemes and the translation words (as in Figure~\ref{entropyexample}) are used to cluster (segment) together neighbor phonemes whose alignment distribution peaks at the same word. The final speech segmentation 
is evaluated using the \textit{Zero Resource Challenge}\footnote{Available at \url{http://zerospeech.com/2017}.} (ZRC) 2017 evaluation suite (track 2).\footnote{
We increment over \cite{godard2018unsupervised} by removing silence labels before training, and using them for segmentation. This results in slightly better scores.}
\subsection{Parallel Speech Corpora}\label{corpora}
The  parallel speech corpora used in this work are the English-French 
(EN-FR)~\cite{kocabiyikoglu2018augmenting}
and the Mboshi-French 
(MB-FR)~\cite{godard2017very} parallel corpora. 
EN-FR corpus is a 33,192 sentences multilingual extension from  \textit{librispeech} \cite{PanayotovCPK15}, 
with English audio books automatically aligned to French translations. MB-FR is a 5,130 sentences corpus from the language documentation process of Mboshi~(Bantu C25), an endangered language spoken in Congo-Brazzaville.
Thus, while the former corpus presents larger vocabulary and longer sentences, the latter presents a more tailored environment, with short sentences and simpler vocabulary.
In order to provide a fair comparison, as well as to study the impact of corpus size, the EN-FR corpus was also down-sampled to 5K utterances (to the exact same size than the MB-FR corpus). Sub-sampling was conducted preserving the average number of tokens per sentence, shown in 
Table~\ref{corpora_tab1}.
\begin{table*}
\centering
\caption{Statistics of the three source-target data sets.}
\begin{tabular}{c|cc|cc|cc|cc}
\toprule
                                          & \multicolumn{2}{c|}{\textbf{\#types}} & \multicolumn{2}{c|}{\textbf{\#tokens}} & \multicolumn{2}{c|}{\textbf{average( token length)}} & \multicolumn{2}{c}{\textbf{average( \#tokens / sentence)}} \\ 
\multicolumn{1}{c|}{\textbf{corpus}}      & \textbf{source}   & \textbf{target}   & \textbf{source}    & \textbf{target}   & \textbf{source}        & \textbf{target}       & \textbf{source}            & \textbf{target}            \\ \hline
\multicolumn{1}{c|}{\textbf{EN-FR (33k)}} & 21,083            & 33,135            & 381,044            & 467,475           & 4.37                   & 4.57                  & 11.48                      & 14.08                      \\ 
\textbf{EN-FR (5k)}                       & 8,740             & 12,226            & 59,090             & 72,670            & 4.38                   & 4.57                  & 11.52                      & 14.17                      \\ 
\textbf{MB-FR (5k)}                       & 6,633             & 5,162             & 30,556             & 42,715            & 4.18                   & 4.39                  & 5.96                       & 8.33 \\ \bottomrule                 
\end{tabular}

\label{corpora_tab1}
\end{table*}

\subsection{Introducing Average Normalized Entropy (ANE)}
In this paper, we focus on studying the soft-alignment probability matrices resulting from the learning of S2S 
models for the UWS task. 
To assess the overall quality of these  matrices without having gold alignment information, we introduce Average Normalized Entropy (ANE).


\begin{figure}
\centering
\includegraphics[scale=0.17]{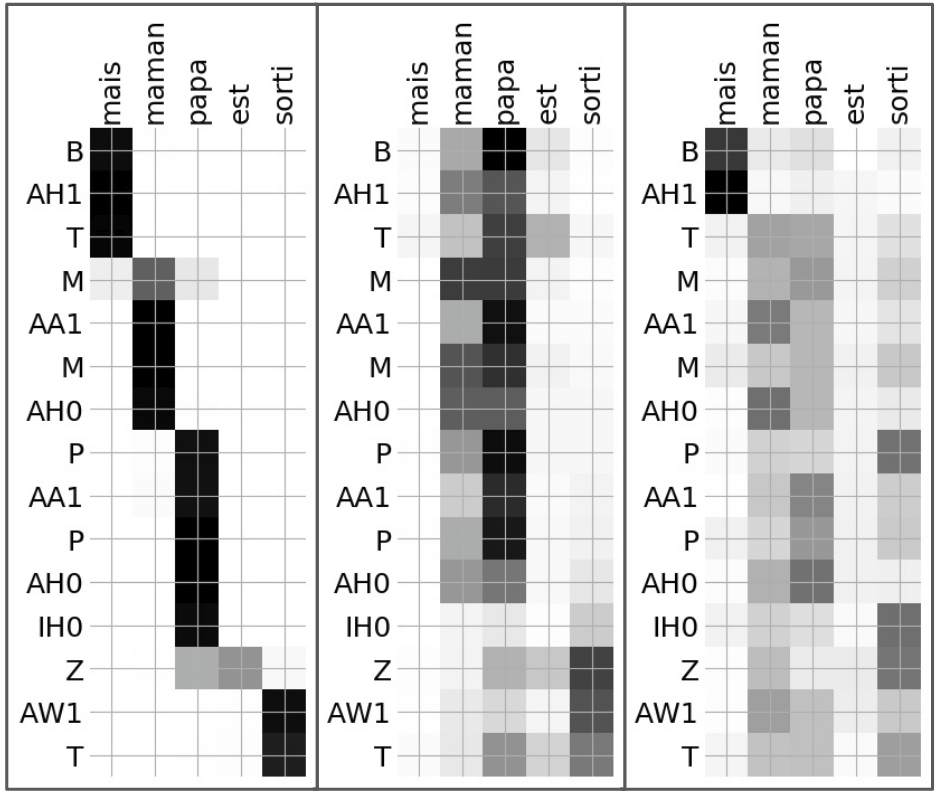}
    \caption{Soft-alignment probability matrices from the UWS task. ANE values (from left to right) are 0.11, 0.64 and 0.83. The gold segmentation is ``BAH1T MAA1MAH0 PAA1PAH0 IH0Z AW1T'', which corresponds to the English sentence ``But mama, papa is out''.}
\label{entropyexample}
\end{figure}

\noindent\textbf{Definition:} 
Given the source and target pair $(\src,\trg)$ of lengths $\lsrc$ and $\ltrg$ respectively,
for every phone $t_i$, the normalized entropy (NE) is computed considering all possible words in $s$ (Equation~\ref{normentropy}), where $P(t_i,s_j)$ is the alignment probability between the phone $t_i$ and the word $s_j$ (a cell in the matrix). 
The ANE for a sentence is then defined by the arithmetic mean over the resulting NE for every phone from the sequence $t$ (Equation~\ref{avgentropy}). 
\begin{equation}\label{normentropy}
    NE(t_i, s) = - 
    \sum_{j=1}^{\lsrc} P(t_{i}, s_{j})\cdot \log_{\lsrc}(P(t_{i},s_{j}))
\end{equation}
\begin{equation}\label{avgentropy}
    ANE(t,s) = \frac{\sum_{i=1}^{\ltrg}NE(t_i, s)}{\ltrg}
\end{equation}
From this definition, we can derive ANE for different granularities (sub or supra-sentential) by accumulating its value for the full corpus, for a single type or for a single token.
\textit{Corpus ANE} will be used to  summarize the overall performance of a S2S model on a specific corpus.
\textit{Token ANE} 
extends  ANE to tokens
by averaging NE for 
all phonemes from a single (discovered) token.
\textit{Type ANE} results from averaging the ANE for every token instance of a discovered type. Finally, \textit{Alignment ANE} is the result of averaging the ANE for every discovered \textit{(type, translation word)} alignment pair.
Intuition that lower ANEs correspond to better alignments
is exemplified in 
Figure~\ref{entropyexample}. 

\section{Empirical Comparison of S2S Models}\label{sec:models}
We compare three NMT models $\S$\ref{bah}, $\S$\ref{transformer}, $\S$\ref{pervasive}) for UWS, focusing on their ability of aligning words (French) with phonemes (English or Mboshi) in medium-low resource settings. The results,  an analysis of the impact of data size and quality, and the correlation between intrinsic (ANE) and extrinsic (boundary F-score) metrics are presented in $\S$\ref{task1exp}. The application of ANE for type discovery in low-resource settings is presented in $\S$\ref{sec:type}.

\subsection{RNN: Attention-based Encoder-Decoder}\label{bah}
The classic RNN encoder-decoder model
~\cite{bahdanau2014neural} 
connects a bidirectional encoder with an unidirectional decoder by the use of an \textit{alignment module}. The RNN encoder learns annotations for every source token, and these are weighted by the alignment module for the generation of every target token.  Weights are defined as \textit{context vectors}, since they capture the importance of every source token for the generation of each target token. 


\noindent\textbf{Attention mechanism:} 
a context vector for a decoder step $t$ is computed using the set of source annotations $H$ and the last state of the decoder network (translation context). The attention is the result of the weighted sum of the source annotations $H$ (with $H = h_1,..., h_A$) and their $\alpha$  probabilities (\ref{seq2seq-attention}) obtained through a feed-forward network $align$ (\ref{seq2seq-attention2}). 
\begin{equation}\label{seq2seq-attention}
c_t = 
Att(H, s_{t-1})= \sum_{i=1}^{A}\alpha_i^t h_i
\end{equation}
\begin{equation}\label{seq2seq-attention2}
\alpha_i^t = {\rm softmax}(align(h_i, s_{t-1})) 
\end{equation}
\subsection{Transformer}\label{transformer}
Transformer~\cite{vaswani2017attention} is a fully attentional S2S 
architecture, 
which has obtained 
state-of-the-art results for several NMT 
shared tasks.
It replaces the use of sequential cell units (such as LSTM) by Multi-Head Attention (MHA) operations, which make the architecture considerably faster. 
Both encoder and decoder networks are 
stacked layers sets that receive source and target sequences, embedded and concatenated with positional encoding. 
An encoder layer 
is made of two sub-layers: a \textit{Self-Attention} MHA 
and a feed-forward. 
A decoder layer 
is made of three sub-layers: a \textit{masked Self-Attention} MHA  
(no access to subsequent positions); an 
\textit{Encoder-Decoder} MHA 
(operation over the 
encoder stack's final output 
and the decoder self-attention output); and a 
feed-forward sub-layer.
Dropout and residual connections are applied between all sub-layers. 
Final output probabilities are generated by applying a linear projection over the decoder stack's output, followed by
a softmax operation. 

\noindent\textbf{Multi-head attention mechanism:}
attention 
is seen as a mapping problem: given a pair of key-value vectors and a query vector, the task is the computation of the weighted sum of the given values (output). In this setup, weights are learned by compatibility functions between key-query pairs (of dimension $d_k$). For a given query (Q), keys (K) and values (V) set, the \textit{Scaled Dot-Product (SDP) Attention} function is computed as: 
\bea
    Att(V, K, Q) = softmax(\frac{QK^T}{\sqrt{d_k}})V
\eea
In practice, several \textit{attentions} are computed for a given QKV set. The QKV set is first projected into $h$ different spaces (multiple heads), where the SDP attention is computed in parallel. Resulting values for all heads are then concatenated and once again projected, yielding the layer's output. (\ref{transformer-multihead1}) and (\ref{transformer-multihead2}) illustrate the process, in which H is the set of $h$ heads ($H = h_1, ..., h_h$) and $f$ is a linear projection.
\textbf{Self-Attention} defines the case where query and values come from same source (
learning compatibility functions within the same sequence of elements).

\begin{equation}\label{transformer-multihead1}
MultiHead(V, K, Q) = f(Concat(H))
\end{equation}

\begin{equation}\label{transformer-multihead2}
h_i = Att(f_i(V), f_i(K), f_i(Q))
\end{equation}

\subsection{CNN: Pervasive Attention}\label{pervasive}
Different from the previous models, which are based on encoder-decoder structures interfaced 
by attention mechanisms, this approach relies on a single 2D CNN across both sequences (no separate coding stages)~\cite{elbayad2018pervasive}.
Using masked convolutions, an auto-regressive model predicts  the next output symbol based on a joint representation of both  input and partial output sequences.
Given a source-target pair $(\src,\trg)$ of lengths $\lsrc$ and $\ltrg$ respectively, tokens are first embed in $d_s$ and $d_t$ dimensional spaces via look-up tables.
Token embeddings $\{x_1,\ldots, x_\lsrc\}$ and $\{y_1,\ldots,y_\ltrg\}$ are then concatenated to form a 3D tensor $X\in\R^{\ltrg\times\lsrc\times f_0}$, with $f_0=d_t+d_s$, where:
\bea
    X_{ij} = [y_i \;\; x_j]
\eea
Each convolutional layer $l\in\{1,\ldots,L\}$ of the model produces a tensor $H_l$ of size $\ltrg\!\times\!\lsrc\!\times\!f_l$, 
where $f_l$ is the number of output channels for that layer.
To compute a distribution over the tokens in the output vocabulary, the second dimension of the tensor is used. This dimension is of variable length (given by the input sequence) and it is collapsed by max or average pooling  to obtain the tensor $H_L^\text{Pool}$ of size $\ltrg\!\times\!f_L$.
Finally, $1\!\times\!1$ convolution followed by a softmax operation are applied, resulting in the distribution over the target vocabulary for the next output token.

\noindent\textbf{Attention mechanism:} 
joint encoding acts as an attention-like mechanism, since individual source elements are re-encoded as the output is generated. 
The self-attention approach of~\cite{lin17iclr} is applied. It computes the attention weight tensor $\alpha$, of size $\ltrg\times \lsrc$,  from the last activation tensor $H_L$, to pool the elements of the same tensor along the source dimension, as follows:
\bea
     \alpha  & = &\softmax\left(W_1 \tanh\left(H_L  W_2\right)\right) \\
    H_L^\text{Att} & =&  \alpha H_L.
\eea
where $W_1\in\R^{f_a}$ and $W_2\in\R^{f_a\times f_L}$ are weight tensors that
map the $f_L$ dimensional features in $H_L$ to the attention weights via an $f_a$ dimensional intermediate representation.




\subsection{Comparing S2S Architectures}\label{task1exp}

For each 
S2S architecture, and each of the three corpora, 
we train five models~(runs) with different initialization seeds.\footnote{RNN, CNN and Transformer implementations from~\cite{berard2016listen,elbayad2018pervasive,ott2019fairseq} respectively.} 
Before segmenting, we
average the produced matrices from the five different runs as in~\cite{godard2018unsupervised}. 
Evaluation is done in a \textit{bilingual} segmentation condition that corresponds to the real UWS task
. In addition, we also perform segmentation in a \textit{monolingual} condition, where a phoneme sequence 
 is segmented with regards to the corresponding word sequence (transcription) in the same language (hence monolingual).\footnote{This task can be seen as an automatic extraction of a pronunciation lexicon from parallel words/phonemes sequences.}
Our 
networks are optimized for the monolingual task.
Across all architectures, we use embeddings of size 64 and batch size of 32 (5K data set), or embeddings of size 128 and batch size of 64 (33K data set). Dropout of 0.5 and early-stopping 
procedure are applied in all cases.
RNN models have only one layer, a bi-directional encoder, and cell size equal to the embedding size, as in~\cite{godard2018unsupervised}. CNN models use the hyper-parameters from~\cite{elbayad2018pervasive} with only 3 layers (5K data set), or 6 (33K data set), and kernel size of 3. 
Transformer models were optimized starting from the original hyper-parameters of \cite{vaswani2017attention}. Best results (among 50 setups) were achieved using 2  
heads, 3 layers (encoder and decoder), warm-up of 5K 
steps, and using cross-entropy loss without label-smoothing. Finally, for selecting which head to use for UWS, we experimented using the last layer's averaged heads,  
or by selecting the head with minimum corpus ANE. While the results were not  significantly different,
we kept the ANE selection.

\subsubsection{Unsupervised Word Segmentation Results}
The word boundary F-scores\footnote{For CNN and RNN, average standard deviation for the bilingual task 
is of less than 0.8\%. For Transformer, it is almost 4\%.} for the task of UWS from phoneme sequence (in Mboshi or English) are presented in Table~\ref{boundarytable}, with monolingual results shown for information only (topline).
Surprisingly, RNN models outperform the more recent (CNN and Transformer) approaches. 
One possible explanation is the lower number of parameters (for a 5K setup, in average 700K parameters are trained, while CNN needs an additional 30.79\% and Transformer 5.31\%). However, for 33K setups, CNNs actually need 30\% less parameters than RNNs, but still perform worse. 
Transformer's low performance could be due to the use of  several heads ``distributing'' alignment information across different matrices. Nonetheless, we evaluated averaged heads and single-head models, and these resulted in significant decreases in performance. 
This suggests that this architecture may not need to learn explicit alignment to translate, but instead it could be capturing different kinds of linguistic information, 
as discussed in the original paper and in its examples~\cite{vaswani2017attention}. 
Also, on the decoder side,  the behavior of the self-attention mechanism on phoneme units is unclear and under-studied so far. For the encoder, Voita et al.~\cite{voita2019analyzing} performed after-training encoder head removal based on \textit{head confidence}, 
showing that after initial training, most heads were not necessary for maintaining translation performance.
Hence, we find the Multi-head mechanism interpretation challenging, and maybe not suitable for a direct word segmentation application, such as our method.

As in \cite{pierre2019thesis}, our best UWS method (RNN) for the bilingual task does not reach the performance level of a strong  Bayesian baseline~\cite{Goldwater09bayesian} with F-scores of 89.80 (EN33K), 87.93 (EN5K), and 77.00 (MB5K). 
However, even if we only evaluate word segmentation performance, our neural approaches learn to segment \textit{and} align, whereas 
this baseline only learns to segment. 
Section \ref{sec:type} will leverage those alignments for a type discovery task useful in language documentation.

The Pearson's $\rho$ correlation coefficients between ANE and boundary F-scores for all mono and bilingual runs of all corpora ($N=30$) 
are $-0.98$ (RNN), $-0.97$ (CNN), and $-0,66$ (Transformer), with  p-values smaller than $10^{-5}$.
These strong negative correlations confirm our hypothesis that lower ANEs correspond to sharper and better 
alignments. 

\begin{table}
\caption{Boundary F-scores for the UWS task.} 
\centering
\begin{tabular}{ll|c|c}
\toprule
                                                     & \multicolumn{1}{c|}{\textbf{}} & \textbf{Bilingual} & \textbf{Monolingual} \\ \midrule
\multicolumn{1}{c}{\multirow{3}{*}{\textbf{EN 33K}}} & \textbf{RNN}                   & 77.10              & 99.80                \\
\multicolumn{1}{c}{}                                 & \textbf{CNN}                   & 71.30              & 98.60                \\
\multicolumn{1}{c}{}                                 & \textbf{Transformer}           & 52.70              & 94.90                \\ \hline
\multirow{3}{*}{\textbf{EN 5K}}                      & \textbf{RNN}                   & 70.40              & 99.30                \\
                                                     & \textbf{CNN}                   & 55.90              & 98.80                \\
                                                     & \textbf{Transformer}           & 52.50              & 80.90                \\ \hline
\multirow{3}{*}{\textbf{MB 5K}}                      & \textbf{RNN}                   & 74.00              & 92.50                \\
                                                     & \textbf{CNN}                   & 68.20              & 89.80                \\
                                                     & \textbf{Transformer}           & 66.40              & 83.50 \\\bottomrule              
\end{tabular}
\label{boundarytable}
\end{table}

\subsubsection{Impact of Data Size and Quality}\label{dataimpact}



EN33K and EN5K results of Table~\ref{boundarytable} allow us to analyze the impact of data size on the S2S models. For the bilingual task, RNN performance drops by 7\% on average, whereas performance drop is bigger for CNN (14-15\%). Transformer performs poorly in both cases, and increasing data size from 5K to 33K seems to help only for a trivial task (see \textit{monolingual} results).

The EN5K and MB5K results of Table~\ref{boundarytable} reflect the impact of language pairs on the S2S models. 
We know from~\cite{fourtassi2013english, rialland2015dropping} that English should be easier to segment than Mboshi, and this was confirmed by both \textit{dpseg} and \textit{monolingual} results. 
However, this trend is not confirmed in the \textit{bilingual} task, where the quality of the (sentence aligned) parallel corpus seems to 
have more impact
(higher boundary F-scores for MB5K than for EN5K for all S2S models).
As shown in Table~\ref{corpora_tab1}, MB-FR corpus has shorter sentences and smaller lexicon diversity, while EN-FR is made of automatically aligned books (noisy alignments), what may explain our experimental results.



\subsection{Type Discovery in Low-Resource Settings}

\begin{table}
\caption{Type retrieval results (RNN) using ANE for keeping most confident  (type, translation) pairs. For instance, $ANE=0.4$ means all discovered types have $ANE\leq0.4$. 
}
\begin{tabular}{r|ccc|ccc}
\toprule
    & \multicolumn{3}{c|}{\textbf{EN 5K}} & \multicolumn{3}{c}{\textbf{MB 5K}} \\
ANE & P       & R      & F      & P       & R      & F      \\\midrule
0.1 & 70.97   & 0.50   & 1.00   & 72.13   & 0.57   & 1.12   \\
0.2 & 55.43   & 3.85   & 7.20   & 49.02   & 2.89   & 5.46   \\
0.3 & 44.99   & 12.51  & 19.58  & 38.18   & 8.14   & 13.41  \\
\textbf{0.4} & 32.81   & 21.76  & \textbf{26.17}  & 32.63   & 16.61  & 22.01  \\
0.5 & 23.37   & 28.17  & 25.54  & 27.93   & 23.44  & 25.49  \\
\textbf{0.6} & 18.54   & 32.41  & 23.59  & 24.73   & 27.61  & \textbf{26.09}  \\
0.7 & 16.23   & 34.34  & 22.04  & 23.00   & 30.12  & 26.08  \\
0.8 & 15.21   & 35.16  & 21.23  & 22.17   & 30.95  & 25.84  \\
0.9 & 15.01   & 35.31  & 21.06  & 22.06   & 31.05  & 25.80  \\
\textbf{all} & 15.01   & 35.34  & 21.07  & 22.06   & 31.05  & 25.80 \\\bottomrule
\end{tabular}
\label{table:typeretrieval}
\end{table}

\label{sec:type}

We investigate the use of \textbf{Alignment ANE} as a confidence measure. From the RNN models, 
we extract and rank the discovered types by their 
ANE, and examine if 
it can be used to separate true words in the discovered vocabulary from the rest. The  
results for low-resource scenarios (only 5K) in 
Table~\ref{table:typeretrieval} suggest that  
low ANE 
corresponds to the portion of the discovered vocabulary the network is \textit{confident} about, and these are, in most of the cases, true discovered lexical items (first row, $P\geq70\%$).\footnote{Type ANE for the retrieval task was also investigated, and results were positive, but slightly worse than the ones from Alignment ANE.} 
As we keep higher Alignment ANE values, we increase recall but loose precision. 
This suggests that, in a documentation scenario, ANE could be used as a confidence measure by a linguist to extract a list of types with higher precision, without having to pass through all the discovered vocabulary. 
Moreover, as exemplified for EN5K in Table~\ref{anetop10}, we also 
retrieve aligned information (translation candidates) for the generated lexicon.


\begin{table}
\caption{Top 5 low and high ANE ranking for the discovered types (EN5K), with gold transcription and aligned information between parentheses (respectively). ``INV'' means incorrect type.
}
\centering
\begin{tabular}{cl|l}
\toprule
\multicolumn{1}{l}{\textbf{ }} & \multicolumn{1}{c}{\textbf{Top Low ANE}} & \multicolumn{1}{|c}{\textbf{Top High ANE}} \\ \midrule
\textbf{1}           & SER1 (sir, $EOS\_token$) & AH0 (a, convenable) \\
\textbf{2}           & HHAH1SH (hush, chut) & IH1 (INV, ah) \\
\textbf{3}           & FIH1SHER0 (fisher, fisher) & D (INV, riant) \\
\textbf{4}           & KLER1K (clerk, clerc) & N (INV, obéit) \\
\textbf{5}           & KIH1S (kiss, embrasse) & YUW1 (you, diable) \\\bottomrule
\end{tabular}
\label{anetop10}
\end{table}

\section{Conclusions}
\label{sec:conclusion}

We presented an empirical evaluation of different architectures (RNN, CNN and Transformer) with respect to their capacity to align word sequences in a source language with phoneme sequences in a target language, inferring 
from it 
word segmentation on the target side (UWS task).\footnote{Pointers for corpora, parameters and implementations available at \url{https://gitlab.com/mzboito/attention_study}} Although RNNs have been outperformed by CNN and Transformer-based 
models for machine translation, for UWS these architectures are still more robust in low-resource scenarios, and present the best segmentation results. 
We also introduced ANE, 
an intrinsic measure of alignment quality of S2S 
models.  
Accumulating it 
over the discovered alignments
, we showed 
it can be used as a confidence measure to 
select 
true words, 
increasing Type F-scores.

\bibliographystyle{IEEEtran}
\bibliography{template}
\end{document}